\pdfoutput=1

\documentclass[11pt]{article}

\usepackage{emnlp2023}

\usepackage{times}
\usepackage{latexsym}
\usepackage{amssymb}
\usepackage{amsmath}
\usepackage{booktabs}
\usepackage{cleveref}
\usepackage{derivative}

\usepackage{graphicx}
\usepackage{tabularx}
\usepackage{algorithm}
\usepackage{algorithmic}
\usepackage{multirow}
\usepackage{float}
\usepackage{footnote}
\makesavenoteenv{tabular}
\usepackage{enumitem}
\usepackage{listings}
\usepackage{todonotes} 
\usepackage[T1]{fontenc}

\usepackage[utf8]{inputenc}

\usepackage{microtype}
\usepackage{xcolor}
\definecolor{ForestGreen}{RGB}{34,139,34}

\newcommand{\que}[1]{\todo[inline,color=green!20!white]{#1}}

\newcommand{\ans}[1]{\todo[inline,color=purple!20!white]{#1}}

\crefformat{footnote}{#2\footnotemark[#1]#3}

\title{Noise Contrastive Estimation-based Matching Framework for Low-Resource Security Attack Pattern Recognition}

\author{Tu Nguyen,
  Nedim  \v{S}rndi\'{c},
  Alexander Neth \\
  Huawei R\&D Munich \\
  \texttt{\{tu.nguyen, nedim.srndic, alexander.neth\}@huawei.com}
}

\begin{document}
\maketitle
\begin{abstract}
Tactics, Techniques and Procedures (TTPs) represent sophisticated \textit{attack patterns} in the cybersecurity domain, described encyclopedically in textual knowledge bases. Identifying TTPs in cybersecurity writing, often called \emph{TTP mapping}, is an important and challenging task. Conventional learning approaches often target the problem in the classical multi-class or multi-label classification setting. 
This setting hinders the learning ability of the model due to a large number of classes (i.e., TTPs), the inevitable skewness of the label distribution and the complex \textit{hierarchical} structure of the label space. We formulate the problem in a different learning paradigm, where the assignment of a text to a TTP label is decided by the direct semantic similarity between the two, thus reducing the complexity of competing solely over the large labeling space. 
To that end, we propose a neural matching architecture with a sampling-based learn-to-compare mechanism and two complementary sampled objectives: $\alpha$-balanced NCE controls the collective mass, but not the internal ordering, of sampled negatives; asymmetric focusing controls individual comparisons under incomplete annotation.
\end{abstract}

\section{Introduction and Background}




Cyber Threat Intelligence (CTI), an essential pillar of cybersecurity, involves collecting and analyzing information on cyber threats, including threat actors, their campaigns, and malware, helping timely threat detection and defense efforts. Textual threat reports or blogs are considered a important source of CTI, where security vendors diligently investigate and promptly detail intricate attacks. A key sub-task in extracting CTI from these textual sources involves the identification of Tactics, Techniques, and Procedures (TTP) of the threat actors, i.e. comprehending descriptions of low-level, complex threat actions and connecting them to standardized attack patterns. One of the popular standard knowledge frameworks widely adopted in the CTI community is MITRE ATT\&CK~\cite{MitreAttackPhilosophy}. Within this framework, a technique represents a specific method used to achieve an objective, with its corresponding tactics and sub-techniques covering broader strategies and variations. Fig.~\ref{fig:attack_text_example1} illustrates an example of a text in a threat report, which indicates two attack patterns, among others, i.e., (1) the use of a malicious email attachment to take control of a victim's system (T1566~\footnote{\href{https://attack.mitre.org/techniques/T1566/}{{\nolinkurl{attack.mitre.org/techniques/T1566}}}}), and (2) encrypting data on the victim's system, presumably for ransom demands (T1486~\footnote{\href{https://attack.mitre.org/techniques/T1486/}{{\nolinkurl{attack.mitre.org/techniques/T1486}}}}). 
\setlength\itemsep{-1em}
\begin{figure}[t]
\centering
\begin{lstlisting}
[...] We witnessed that the botnet was spread via mass phishing, using a VB-scripted Excel attachment to download the second stage from xx.warez22.info. The same domain was used for C&C via HTTP. The botnet distributed a file encryption module we named VBenc. [...]
\end{lstlisting}
\caption{A fictional attack described in typical cybersecurity threat report writing style.}
\label{fig:attack_text_example1}
\end{figure}
\setlength\itemsep{-1em}



As of~2024, there are over~600 techniques, together with~14 high-level tactics described in MITRE ATT\&CK. In its ontology, a technique is associated to at least one tactic (e.g., the technique ``Hijack Execution Flow'' is listed under three distinct tactics: Persistence, Privilege Escalation and Defense Evasion) and may have several sub-techniques. Mining techniques from CTI reports poses significant challenges due to several factors. Firstly, the large number of techniques, coupled with their diverse nature, intricate inter-dependencies, and hierarchical structure, renders the task complex and laborious. Secondly, the analysis of CTI reports necessitates the expertise of security professionals. The reports focus on delineating low-level threat actions rather than explicitly mentioning the associated techniques and tactics. Consequently, extracting relevant techniques and tactics from these reports requires diligent inference by the reader. Employing an automated approach to TTP mapping presents inherent challenges. One major hurdle is the \textit{low-resource} nature of the task, due to the limited availability of labeled data and the extensive label space. Moreover, the presence of long-tail infrequent TTPs adds complexity to the learning process.  


Due to these challenges, TTP mapping has not been fully solved in related work. Most recent works use a classical \textit{document}-level multi-label~\cite{Li_Zheng_Liu_Yang_2019} or \textit{sentence}-level multi-class classification~\cite{9978947, You_Jiang_Jiang_Yang_Liu_Feng_Wang_Li_2022} learning setting. These granularity choices, however, either introduce unneeded complexity of long-form text representation (for \textit{document}-level) or make the task inapplicable to mapping complex TTPs, which often require longer text (for \textit{sentence}-level). Moreover, the main learning issues in these settings are: (i) the aforementioned problems of label scarcity and long-tailedness, and (ii) the learning complexity costs of the softmax-based learning approaches grow proportionally to the number of classes. In the wider literature i.e., extreme multi-label text classification (XMTC), the problems are addressed by (i) capturing the label correlation and (ii) partitioning and handling the sub-label spaces separately. They are, however, most effective in relatively resource-rich settings, and have drawbacks when applied to~\emph{label-scarce} scenarios, as the signal-to-noise ratio increases~\cite{Bamler_Mandt_2020}. 
In the multi-label context, learning is greatly affected, additionally, by the frequent presence of \textit{missing} labels, which is a common trait observed in human-curated datasets.


In this work we propose an alternative learning setting which avoids the direct optimization for discriminating between data points in a large label space. Concretely, we transform the task into a \textit{text matching} problem~\cite{tay2018co,wang2017bilateral}, allowing us to utilize the direct semantic similarity between the \textit{input}-\textit{label} pairs to derive an assignment score used for ranking and thresholding. The framework inherently incorporates an \textit{inductive bias}, encouraging the capture of nuanced similarities even in the presence of limited labeled data, enhancing its ability to generalize to long-tail TTPs. This transformation is achieved by leveraging the \textit{textual profile} of a TTP (i.e., textual \textit{description}~\footnote{ \label{mitrenote}
A technique, its description and procedure examples: \href{https://attack.mitre.org/techniques/T1021/}{\nolinkurl{attack.mitre.org/techniques/T1021/}}} in ATT\&CK), a resource that is often neglected in related work.

\textbf{Label-efficient text matching}: Our approach---dynamic \textit{label-informed} text matching---uses Noise Contrastive Estimation (NCE)~\cite{Gutmann_Hyvarinen_2010} to combine semantic matching with the discriminative pressure of classification. Instead of a fixed label-wise output head, each observed text--label pair is compared with a candidate set that changes throughout training and is drawn from the corpus-level label inventory.

Conventionally, NCEs are used to alleviate computation in very large target spaces. We apply them in a \textbf{moderately sized label space}: too broad for exhaustive comparison at every update, but bounded enough to be progressively explored. Each update yields only a partial ranking; repeated updates progressively broaden label exposure. We introduce two complementary loss modifications. $\alpha$-balanced NCE controls candidate-set mass without changing its internal ordering, whereas asymmetric focusing controls individual pair influence under missing-label noise. This set-level/pair-level design is our loss contribution and connects finite-space coverage with incomplete supervision.

To this end, we summarize our contributions:
\setlength\itemsep{-0.5em}
\begin{itemize}
\setlength\itemsep{-0.5em}
\item We formally redefine the challenging task of TTP mapping as a \textit{paragraph-level} \textit{hierarchical} multi-label text classification problem and propose a new learning paradigm that works effectively on the nature of the task. 
\item We introduce two sampled ranking objectives for dynamic label-informed matching: set-level $\alpha$-balancing and pair-level asymmetric focusing. We also characterize how a moderate label space trades per-update cost for cumulative coverage, and exploit the label hierarchy through multi-task learning.
\item We curate and publicize an expert-annotated dataset that emphasizes on the multi-label nature, with approximately two times more labels per sample than existing datasets.
\item Lastly, we conduct extensive experiments to prove our learning methods outperform strong baselines across real-world datasets. 
\end{itemize}

\section{Related Work}
\setlength\itemsep{-1em}
\textbf{TTP Mapping and CTI Extraction}
\setlength\itemsep{-1em}
Several works target TTP mapping on the \emph{document level}. ~\cite{Husari_Al-Shaer_Ahmed_Chu_Niu_2017} used a probabilistic relevance framework (Okapi BM25) to quantify the similarity between \textit{BoW} representations of TTPs and the target text. However, this approach is limited to the oversimplified vocabulary of threat actions within an \textit{ad-hoc} ontology. \citet{ayoade2018automated,niakanlahiji2018natural} used a TF-IDF-based document representation and leveraged classical (i.e., tree-based, margin-based) ML for (multi-label) classification. \citet{Li_Zheng_Liu_Yang_2019} used latent semantic analysis to extract topics from target articles, and compared the topic vectors with the TF-IDF vectors of ATT\&CK description pages to obtain cosine similarity. They used the similarity vectors with Na\"ive Bayes and decision trees to classify TTPs. However, the choice of document-level granularity introduces additional unneeded complexity of long-form text representation.
Recent works leverage transformers for \textit{sentence-level} text representation learning~\cite{9978947,You_Jiang_Jiang_Yang_Liu_Feng_Wang_Li_2022}, using the encoded representation in the multi-class classification setting. However, with limited available data, they restrict the task to only a small number of TTPs.

\setlength\itemsep{-1em}
\textbf{Extreme Multi-label Text Classification}.
\setlength\itemsep{-1em}
XMTC, or generally extreme multi-label classification is a line of research targeting extremely large label spaces, e.g., product categorization in e-commerce or web page categorization. The main challenges for XMTC are computational efficiency and data skewness. Common techniques for XMTC are tree-based~\cite{you2019attentionxml,jasinska2020probabilistic,wydmuch2018no}, sampling-based~\cite{jiang2021lightxml} and embedding-based~\cite{chang2021extreme} that attempt to partition the label space and thus reduce the computational complexity. However, generally, these methods assume the sufficient availability of supervision and still suffer in the long-tail performance.

\setlength\itemsep{-1em}
\textbf{Matching Networks}.
\setlength\itemsep{-1em}
Deep matching networks have witnessed rapid progress recently, finding applications in various conventional (e.g., retrieval~\cite{wang2017bilateral}) or emerging tasks (e.g., few-shot~\cite{vinyals2016matching} and self-supervised learning~\cite{chen2020simple}). They can be architecturally categorized as \textit{cross}- vs \textit{dual}-encoder networks and can be optimized in tandem with the \textit{triplet}~\cite{schroff2015facenet} or \textit{contrastive loss}~\cite{chopra2005learning}. The former loss considers triplets of examples (anchor, positive, negative) and is \textit{marginal}-based, whereas the latter, broadly referred to as NCE~\cite{Gutmann_Hyvarinen_2010}, utilizes a probabilistic interpretation. Despite demonstrating promising results across various domains and datasets, matching networks necessitate substantial training data. Although the NCE framework partially mitigates this concern, the well-adopted approach by ~\citet{oord2018representation} remains somewhat limited, especially to the \textit{fully-supervised} settings. Our approach overcomes the present constraints of training matching networks in settings where resources are limited, specifically when there is a scarcity of extensive training data. 

\section{Preliminaries and Problem Setup}
\label{sec:pre}

We first introduce notation for label classification and then specialize it to text--label matching.

\textbf{Classification.} Let $\mathcal{X}$ be the input space and let $\mathcal{L}$ be a finite label set with $M=|\mathcal{L}|$. A parametric classifier assigns a real-valued score
$s_{\theta}:\mathcal{X}\times\mathcal{L}\rightarrow\mathbb{R}$ to every input--label pair. In the single-label setting, the corresponding prediction is
$\widehat{\ell}_{\theta}(x)\in\arg\max_{\ell\in\mathcal{L}}s_{\theta}(x,\ell)$. In the multi-label setting, all $M$ scores are ranked or thresholded, and the target is represented by a multi-hot vector in $\{0,1\}^{M}$.

\textbf{Matching.} Each label $\ell\in\mathcal{L}$ has a textual profile $r(\ell)$ in a text space $\mathcal{T}$. We learn a differentiable matching score
$g_{\theta}:\mathcal{X}\times\mathcal{T}\rightarrow\mathbb{R}$, with parameters $\theta\in\mathbb{R}^{P}$. The text input and the label profile need not be related by an invertible projection; their encoders map them into a common latent space in which $g_{\theta}$ is evaluated.

\textbf{Cross-entropy and sampled contrastive learning.}
For a single-label softmax model,
\begin{equation}
\begin{aligned}
 p_{\theta}(\ell\mid x)
 &=\frac{\exp s_{\theta}(x,\ell)}
 {\sum_{\ell'\in\mathcal{L}}\exp s_{\theta}(x,\ell')},\\
 \mathcal{J}_{\mathrm{CE}}(\theta)
 &=\mathbb{E}_{(x,\ell)\sim p_{\mathrm{data}}}
 \left[-\log p_{\theta}(\ell\mid x)\right].
\end{aligned}
\label{eq:ce}
\end{equation}
Computing the denominator is expensive when $M$ is large. Sampled contrastive objectives replace the exhaustive comparison with one or more observed positive labels and $K$ labels drawn from a noise distribution $q$. The resulting score is used for matching and ranking; unless separately calibrated, it should not be interpreted as a calibrated posterior over the complete label set.

\textbf{Multi-label classification.}
Let the training set be $\mathcal{D}=\{(x_i,Y_i)\}_{i=1}^{n}$, where $x_i\in\mathcal{X}$ and $Y_i\subseteq\mathcal{L}$ is the set of observed labels for $x_i$. Equivalently, $Y_i$ can be represented by a multi-hot vector $\mathbf{y}_i\in\{0,1\}^{M}$. A conventional classifier learns $f_{\theta}:\mathcal{X}\rightarrow\mathbb{R}^{M}$ and ranks or thresholds its components.

\textbf{Problem reformulation.}
For every input $x_i$ and label $\ell$, define the pairwise supervision
$z_{i\ell}=\mathbf{1}[\ell\in Y_i]$. Our model evaluates the textual profile $r(\ell)$ and learns
\begin{equation}
\begin{aligned}
 g_{i\ell}
 &:=g_{\theta}\bigl(x_i,r(\ell)\bigr)\in\mathbb{R},\\
 p_{\theta}(z_{i\ell}=1\mid x_i,r(\ell))
 &=\sigma(g_{i\ell}).
\end{aligned}
\label{eq:pairprob}
\end{equation}
This formulation turns the large-output classification problem into repeated text--label comparisons while preserving the original multi-label supervision.

\section{Methodology}
\setlength\itemsep{-1em}
Here we describe the matching function $g_{\theta}(x,r(\ell))$ and the sampled objectives used to compare candidate TTP labels.
\subsection{Matching Network}
\begin{figure}[h!]
\centering
  \includegraphics[width=.8\columnwidth]{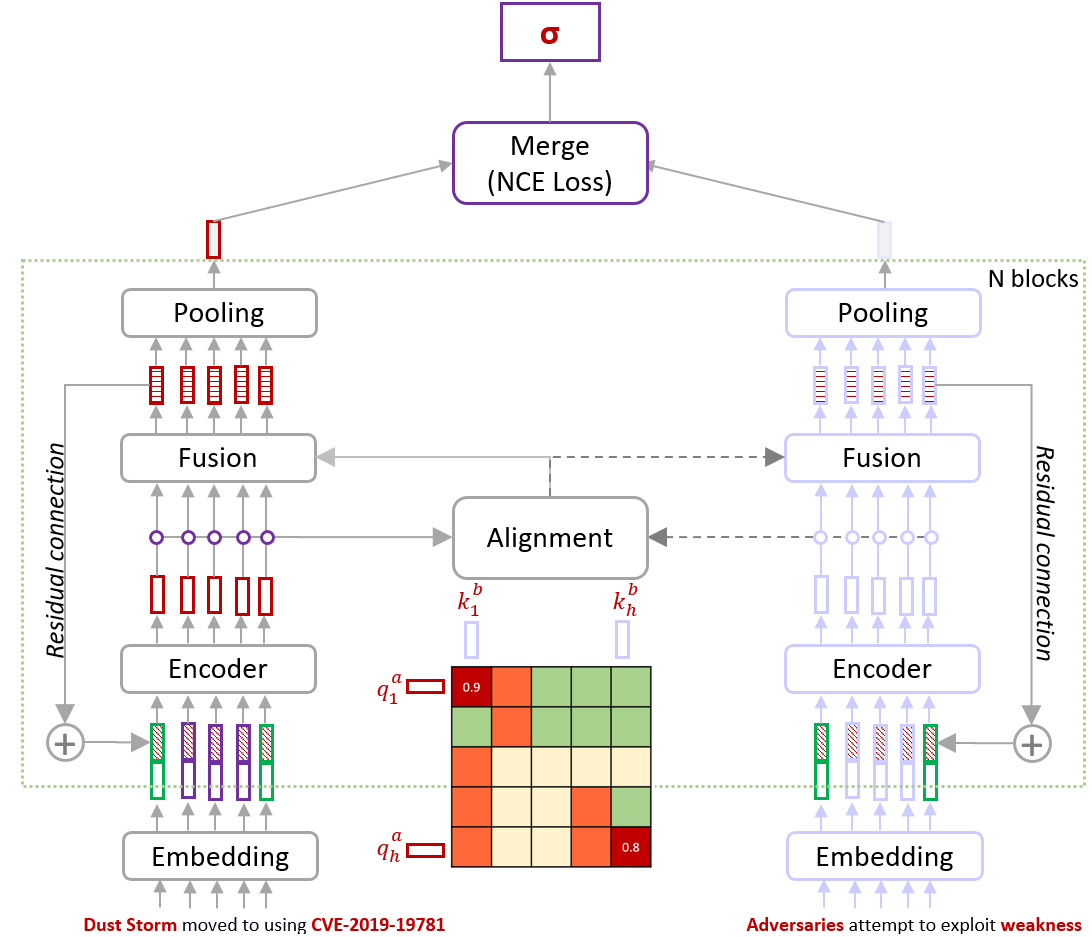}
  \caption{The dual-encoder matching network.}
  \label{fig:bien}
\end{figure}

\setlength\itemsep{-1em}
The architecture of our matching network is built upon the \textit{dual}-encoder framework, which typically employs a Siamese network. This shared network is used for learning the representations of both the target text segment and the TTP \textit{textual profile}. As depicted in Fig.~\ref{fig:bien}, at a high level, our network comprises an embedding component and an alignment component. Each includes specific layers aimed at enhancing the connectivity between the two sub-network sides. Finally, the two sides are merged (e.g., by a dot product) to output a matching logit; a sigmoid maps this logit to the pairwise score in Equation~\ref{eq:pairprob}. We detail the architectural choice for our matching network below.
%

\textbf{Encoder}. The encoder has two modes: (1) \textit{scratch} and (2) \textit{scratch} with a pre-trained transformer (i.e., SecBERT) combined. Scratch indicates that the token embeddings are learnt (with the embedding layer). We then apply a simple CNN on top of the embedding layer. With \textit{scratch} alone, a specialized tokenizer (that respects CTI entities, e.g., URL, vulnerability identifier..) is used. While using together with the transformer, the tokenizer of the transformer is used. For (2), we simply stack the encoded vectors from the two sources together.
 
\textbf{Alignment Network}. Formally, given the input representation of the text-TTP pair as $x_t = (\hat{a}_{1},\ldots,\hat{a}_{l})$ and $y_{ttp} = (\hat{b}_{1},\ldots,\hat{b}_{l})$, the unnormalized attention weights are decomposed into:
$e_{ij} = W^{align}(\hat{a_{i}}) \cdot W^{align}(\hat{b_{j}})$,
whereas $W^{align}$ is a trainable projection matrix, $\cdot$ is the dot product. Then, we derive the normalized weights for each token $a_{i}$ and $b_{j}$, and achieve the corresponding alignment features. Similar to ~\cite{yang2019simple}, we further use the block-based residual architecture with skip connections. Our block consists of the encoder, alignment and fusion layers. The fusion layer does various comparisons of local and aligned representations (i.e., the Hadamard product) and finally fuses the interaction vectors together using the concatenation operator. Then \textit{pooling}, i.e., (non-) weighted average or max-pooling, is applied to attain fixed-length vector representations.
\setlength\itemsep{-1em}
\subsection{Learning to Match and Contrast}
\setlength\itemsep{-1em}
\setlength\itemsep{-1em}
Our efficient learning method avoids exhaustive optimization over the label space. We shift from a fixed multi-label classifier to \textit{dynamic label-informed matching}, where candidate labels are drawn at every step. The ranker, acting simultaneously as a matcher, assigns higher scores to positive pairs and lower scores to candidate pairs. The top-$n$ labels are then selected by ranking and thresholding. We detail the learning mechanism below.

\textbf{\textit{Partial-ranking}-based NCE.}
The purpose of sampling in our setting is to avoid exhaustive ranking over $\mathcal{L}$. Let $(x,\ell^{+})$ be an observed positive pair and draw $K$ candidate negatives $\ell^-_1,\ldots,\ell^-_K$ from a distribution $q_x$ whose support excludes the observed positive labels of $x$. With the pairwise probability in Equation~\ref{eq:pairprob}, define
$p_{+}=p_{\theta}(z=1\mid x,r(\ell^{+}))$ and
$p_j^{-}=p_{\theta}(z=1\mid x,r(\ell_j^{-}))$. A pointwise sampled binary objective is
\begin{equation}
\mathcal{J}_{\mathrm{NCE}}^{\mathrm{local}}(\theta)
=-\mathbb{E}\left[
\log p_{+}+\sum_{j=1}^{K}\log(1-p_j^{-})
\right].
\label{eq:nce}
\end{equation}
Because annotations may be incomplete, a label sampled outside the observed set is a candidate negative rather than a guaranteed semantic negative.

We instead use a sampled ranking objective in which the positive label competes jointly with the $K$ candidates. Write
$g_{+}=g_{\theta}(x,r(\ell^{+}))$ and
$g_{j}^{-}=g_{\theta}(x,r(\ell_j^{-}))$. For $\gamma>0$, define
\begin{equation}
\begin{aligned}
 Z_{\gamma}
 &=\exp(g_{+})+\gamma\sum_{j=1}^{K}\exp(g_j^{-}),\\
 \mathcal{J}_{\alpha\text{-bal}}(\theta)
 &=-\mathbb{E}\left[
 \log\frac{\exp(g_{+})}{Z_{\gamma}}
 \right].
\end{aligned}
\label{eq:infonce}
\end{equation}
For $\gamma=1$, Equation~\ref{eq:infonce} is the usual sampled InfoNCE softmax. When $0<\gamma<1$,
$Z_{\gamma}=\exp(g_{+})+\sum_j\exp(g_j^{-}+\log\gamma)$: every negative receives the same logit shift. Writing $\pi_j^{-}=\gamma\exp(g_j^{-})/Z_{\gamma}$ and $\Pi_{-}=\sum_j\pi_j^{-}$, the ratio $\pi_j^{-}/\Pi_{-}=\exp(g_j^{-})/\sum_k\exp(g_k^{-})$ is independent of $\gamma$, while $\partial\Pi_{-}/\partial\log\gamma=\Pi_{-}(1-\Pi_{-})>0$. Thus, $\gamma$ changes aggregate negative competition without changing within-set ordering: $K$ determines ranking breadth and $\gamma$ its set-level strength. We call this variant \textbf{$\alpha$-balanced NCE}.

\textbf{Asymmetric focusing.}
Uniformly sampled candidates include many easy labels and may also include unannotated positives. Following~\citet{Ridnik_Ben-Baruch_Zamir_Noy_Friedman_Protter_Zelnik-Manor_2021}, we adapt asymmetric focusing from an exhaustive multi-label output to dynamically sampled text--label pairs. Let $\gamma_{+}\geq 0$, $\gamma_{-}>0$, $p=\sigma(g_{\theta}(x,r(\ell)))$, and $m\in[0,1)$ be the negative cutoff. With $p_m=\max(p-m,0)$, the pairwise losses are
\begin{equation}
\begin{aligned}
\ell^{(+)}(p)
  &= -(1-p)^{\gamma_{+}}\log p,\\
\ell^{(-)}(p)
  &= -(p_m)^{\gamma_{-}}\log(1-p_m).
\end{aligned}
\label{eq:asym-pair}
\end{equation}
The sampled asymmetric objective is
\begin{equation}
\mathcal{J}_{\mathrm{asym}}(\theta)
=\mathbb{E}\left[
\ell^{(+)}(p_{+})
+\sum_{j=1}^{K}\ell^{(-)}(p_j^{-})
\right],
\label{eq:asym}
\end{equation}
where $p_{+}=\sigma(g_{+})$ and $p_j^{-}=\sigma(g_j^{-})$. With $\gamma_{-}>\gamma_{+}$, easy negatives are down-weighted more strongly, and $p_j^{-}\leq m$ gives zero loss. The two objectives act at complementary scales: $\alpha$-balanced NCE controls candidate-set competition, whereas asymmetric focusing controls the influence of each sampled pair. This set-level/pair-level decomposition is the loss contribution of our framework. The latter remains a robustness heuristic, not a false-negative detector or a positive--unlabeled consistency guarantee.

Let $\mathcal{D}_{+}=\{(x_i,Y_i,\ell_i^{+}):\ell_i^{+}\in Y_i\}$ and let $\ell_{\mathrm{train}}(g_{+},\mathbf{g}^{-})$ denote the per-example integrand of either Equation~\ref{eq:infonce} or Equation~\ref{eq:asym}. Algorithm~\ref{algo:infonce} gives their common sampled training loop; the corresponding optimization analysis is in Appendix~\ref{apd:conv}.

\begin{algorithm}[t]
\caption{Dynamic label-informed matching}
\label{algo:infonce}
\normalsize
\begin{algorithmic}[1]
\STATE Initialize parameters $\theta$
\FOR{$t=0,\ldots,T-1$}
  \STATE Sample a mini-batch $\mathcal{B}_t\subset\mathcal{D}_{+}$
  \FORALL{$(x_i,Y_i,\ell_i^{+})\in\mathcal{B}_t$}
    \STATE Draw $C_i^{-}\sim q_{x_i}^{K}$ on $\mathcal{L}\setminus Y_i$
    \STATE Evaluate $g_{i,+}$ and $\mathbf{g}_{i}^{-}$ on $\{\ell_i^{+}\}\cup C_i^{-}$
    \STATE $\ell_i\leftarrow\ell_{\mathrm{train}}(g_{i,+},\mathbf{g}_{i}^{-})$
  \ENDFOR
  \STATE $\widehat{\mathcal{J}}_t\leftarrow |\mathcal{B}_t|^{-1}\sum_{i\in\mathcal{B}_t}\ell_i$
  \STATE $\theta\leftarrow\theta-\epsilon_t\nabla_{\theta}\widehat{\mathcal{J}}_t$
\ENDFOR
\STATE \textbf{return} $\theta$
\end{algorithmic}
\end{algorithm}

\subsection{Sampling Strategies}

\textbf{Corpus-level negative sampling.}
In-batch sampling limits the candidate pool to labels represented in the current mini-batch~\cite{Yih_Toutanova_Platt_Meek_2011,Gillick_Kulkarni_Lansing_Presta_Baldridge_Ie_Garcia-Olano_2019}. Corpus-level sampling instead draws labels from the full label set and can therefore expose the model to a broader range of candidates. It does not, however, guarantee that every sampled label is a true negative: incomplete annotations can turn an unobserved positive into a false negative. We treat this as label noise and use the asymmetric weighting in Equation~\ref{eq:asym} as a heuristic mitigation rather than assuming that such errors cancel automatically.

\textbf{Random sampling.}
For an input with observed label set $Y_i$, we use the uniform distribution
$q_{x_i}(\ell)=1/(M-|Y_i|)$ on $\mathcal{L}\setminus Y_i$. Uniform sampling avoids imposing an additional semantic or retrieval bias. Hard-negative samplers are also possible, but can increase the false-negative rate when annotations are incomplete.

\textbf{Moderately sized label space.}
Let $M_i=M-|Y_i|$. The entropy $H(q_{x_i})=\log M_i$ describes the potential breadth of uniform sampling, not the diversity realized in one candidate set. Under independent sampling with replacement, if $U_{i,K,R}$ is the number of distinct labels encountered after $R$ visits to $x_i$, with $K$ draws per visit, then
\begin{equation}
\begin{aligned}
\rho_i(K,R)
&:=\frac{\mathbb{E}[U_{i,K,R}]}{M_i}\\
&=1-\left(1-\frac{1}{M_i}\right)^{KR}
 \approx 1-e^{-KR/M_i}.
\end{aligned}
\label{eq:moderate-coverage}
\end{equation}
Each visit costs $\mathcal{O}(K)$, while expected coverage $1-\delta$ requires approximately
$R_{\delta}\simeq(M_i/K)\log(1/\delta)$ visits. The essential property is temporal: $K\ll M_i$ keeps each update local, while $KR=\Omega(M_i)$ lets diversity accumulate along training. The space is therefore large enough to benefit from sampling, yet small enough for candidate exposure to become progressively global.

\subsection{Hierarchical Multi-label Learning}

In ATT\&CK, TTPs have a hierarchical structure, where different sub-techniques map \textit{many-to-1} to the same technique and techniques map \textit{many-to-many} to tactics. To exploit and encode this structure, we design an \textit{auxiliary} task that predicts the tactics of the textual input, alongside our \textit{matching} task. This auxiliary task is thus also a medium-sized multi-label classification task, and we use the binary cross-entropy loss for the optimization. The two tasks are jointly optimized in a \textit{multi-task} learning manner, where the two losses are linearly combined: $\mathsf{J}_{total} =  \alpha \mathsf{J}_{NCE} + \beta \mathsf{J}_{aux}$, where $\alpha$ and $\beta$ are loss-weighting parameters.


\section{Experiments}

\begin{figure}[tb]
  \centering
  \includegraphics[width=1.\columnwidth]{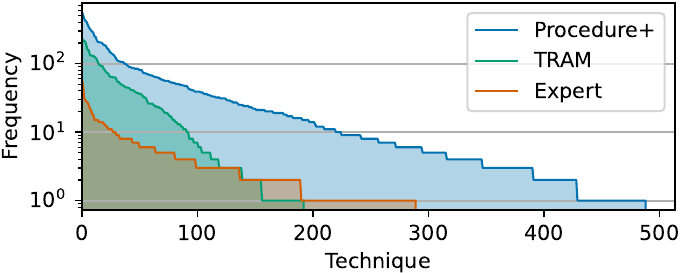}
  \caption{The distributions of the number of samples per technique (TTP) for each dataset.}
  \label{fig:ddist}
\end{figure}

\subsection{Datasets}

We list below the datasets used in our experiments.

\textbf{TRAM.} Largest publicly available manual curated dataset from CTID~\footnote{
  CTID TRAM: \href{https://github.com/center-for-threat-informed-defense/tram}{\nolinkurl{github.com/CTID/TRAM}}}, commonly used in related work.
It comprises mostly short texts, covers only one-third of TTPs with relatively noisy labels, thus appears to have limited application value.

 
\textbf{Procedure+.\footnote{
  \label{datanote} The datasets are publicly shared at \href{https://github.com/tumeteor/ttp-mapping}{\nolinkurl{github.com/TTP-Mapping}} to foster further research.}} \textit{Procedures}: collected from ATT\&CK, where techniques have associated manually curated procedure examples\cref{mitrenote}. Each example is a one-sentence expert-written summary of the implementation of a technique in real-world attacks. \textit{Derived Procedures}: complements an example with a text that aligns to threat report writing style. We look for evidential paragraphs in the references where the summary example is assumedly derived from, using a per-document search engine.

\textbf{Expert.\cref{datanote}} 
Our purposefully crafted dataset closely emulates real-world scenarios, providing a practical setting for TTP extraction. Unlike sentence-focused datasets, ours covers entire paragraphs, thus the annotations are inherently multi-label in nature. Annotated by 5 CTI experts using an in-house tool, our dataset triples text length and increases average labels per sample by approximately 60-80\% compared to TRAM (see Table\ref{tab:datasets}).

In our experiments, the two procedure examples datasets serve as high-quality \textit{pseudo}-datasets, providing additional training examples, as well as valuable benchmarking perspective. Further descriptions of the overall dataset construction processes can be found in Appendix~\ref{sec:dataconst}.
\begin{table}
\caption{Dataset statistics. \emph{S+T} denotes the joint count of techniques and sub-techniques.}
\resizebox{\columnwidth}{!}{%
\begin{tabular}{lrcccc}
\toprule
\centering 

\multirow{2}*{\textbf{Dataset}} & 
\multirow{2}*{\textbf{Texts}} & 
\multirow{2}*{\textbf{S+T}} & 
\textbf{Tech-} & 
\textbf{Avg. \#} & 
\textbf{Avg. \#} \\

&
&
&
\textbf{niques} &
\textbf{Labels} &
\textbf{Tokens} \\

\midrule
\textit{TRAM}               & 4797       & 193   & 132    & 1.16       & 23         \\
\textit{Procedures}         & 11723      & 488   & 180    & 1.00       & 12         \\
\textit{Derived Procedures} & 3519       & 374   & 167    & 1.22       & 65 \\
\textit{Expert}             & 695        & 290   & 151    & \textbf{1.84 }      & \textbf{72}  \\
\bottomrule
\end{tabular}
}
\label{tab:datasets}
\end{table}
\begin{table*}[t]
\caption{Results of all models on~3 datasets. \textit{Procedures+} denotes the combined procedure examples datasets. Bold denotes \textit{best} while underscore signifies  \textit{second-best} performance. Indented (\textit{w/o}) denotes training \textbf{without} the specific option wrt. the preceding model. \textit{Ideal} R$@1$ on the Expert dataset is~0.504. $^{\mathcal{T}}$ uses pre-trained SecBERT.} 
\label{tab:results-all}
\resizebox{\linewidth}{!}{%
\begin{tabular}{clcccccccccccc}
\toprule
\multicolumn{1}{l}{}      &                  & \multicolumn{4}{c}{\textbf{Procedures+}}    & \multicolumn{4}{c}{\textbf{TRAM}}   & \multicolumn{4}{c}{\textbf{Expert}} \\ 
\cmidrule(r){3-6} \cmidrule(r){7-10} \cmidrule(r){11-14}
\multicolumn{1}{l}{}      & \textbf{Methods}          & P$@1$ & R$@1$ & F1$@3$ & MRR$@3$ & P$@1$ & R$@1$ & F1$@3$ & MRR$@3$ & P$@1$ & R$@1$ & F1$@3$ & MRR$@3$ \\
\midrule
\parbox[t]{2mm}{\multirow{3}{*}{\rotatebox[origin=c]{90}{Baseline}}} & TTPDrill (BM25)              & .230   & .227   & .118      & .232    & .250   & .212   & .118     & .205     & .222   & .037   & .008      & .139     \\
                          & Binary Relevance$^{\mathcal{T}}$          & .206   & .579   & .193      & .579     & .236   & .594   & .209      & .594     & .189   & \textbf{.256}   & .085      & .256     \\
                          & \textit{Dynamic} Triplet-loss$^{\mathcal{T}}$ & .339    & .336   & .277      & .432 & .286   & .253   & .277     & . 402    & .449   & .111   & .252      & .525     \\
\midrule
\parbox[t]{2mm}{\multirow{5}{*}{\rotatebox[origin=c]{90}{XMTC}}} 
                          & eXtremeText (Sigmoid)          & .557   & .547   & .371      & .624     & .632   & .594   & .425      & .729     & .407  & .174   & .279      & .485    \\
                          & eXtremeText (PLT)          & .528   & .519   & .336      & .582     & .612   & .578   & .393      & .671     & .344    & .146   & .243      & .411     \\
                          & NAPKINXC             & .578   & .570   & .383      & .661     & .662   & .614   & .453      & .754     & .497   & .199   & .365      & .582    \\
                          & XR-LINEAR & .604    & .595   & .393      & .684     & .674   & .626   & .445      & .757     & .529   & \underline{.215}   & .363      & .600 \\
                          & XR-TRANSFORMER$^{\mathcal{T}}$          & .502   & .494   & .304      & .548     & .540   & .515   & .334      & .595     & .389   & .149   & .239      & .453     \\
\midrule
\parbox[t]{2mm}{\multirow{4}{*}{\rotatebox[origin=c]{90}{Ours}}}   
                          &  InfoNCE$^{\mathcal{T}}$      & .672   & .639   & .442 & .758     & .697   & .577   & .516      & .799     & \underline{.702}   & .175 & \underline{.432}      & \underline{.768}     \\
                          & $@-$balanced$^{\mathcal{T}}$          & \textbf{.760}   & \textbf{.720}   & \underline{.489}      & \underline{.837}     & \underline{.765}   & \underline{.646}  & .\underline{546}     & \underline{.856}     & .693   & .169   & .400      & .762     \\
                          & \hspace{2mm} \textit{w/o} auxiliary     & .604   & .584   & .433      & .719     & .712   & .601   & .521      &  .816   & .693   & .177   & \textbf{.442}      & .773  \\  
                          & \hspace{2mm} \textit{w/o} Transformers    & .646   & .601   & .357      & .772     & .642   & .543   & .547      & .785     & .700   & .173   & .430 & .766  \\    
                          & Asymmetric$^{\mathcal{T}}$       & \underline{.757}   & \underline{.718}   & \textbf{.493}      & \textbf{.838}     & \textbf{.770}   & \textbf{.658}   & \textbf{.555}      & \textbf{.864}     & \textbf{.731}   & .182   & .399      & \textbf{.789}    \\
\bottomrule
\end{tabular}
}
\end{table*}

\begin{table*}[htp]
\caption{\textbf{Technique-level} (resolve sub-techniques to their super-techniques) results, with legend of Table~\ref{tab:results-all} applies.}
\label{tab:results-only-technique} 
\resizebox{\linewidth}{!}{%
\begin{tabular}{clcccccccccccc}
\toprule
\multicolumn{1}{l}{}      &                  & \multicolumn{4}{c}{\textbf{Procedures+}}    & \multicolumn{4}{c}{\textbf{TRAM}}   & \multicolumn{4}{c}{\textbf{Expert}} \\ 
\cmidrule(r){3-6} \cmidrule(r){7-10} \cmidrule(r){11-14}
\multicolumn{1}{l}{}      & \textbf{Methods}          & P$@1$ & R$@1$ & F1$@3$ & MRR$@3$ & P$@1$ & R$@1$ & F1$@3$ & MRR$@3$ & P$@1$ & R$@1$ & F1$@3$ & MRR$@3$ \\
\midrule
\parbox[t]{2mm}{\multirow{3}{*}{\rotatebox[origin=c]{90}{Baseline}}} 
                          & TTPDrill (BM25)        & .294    & .290  & .152   & .297  & .281   & .271   & .161  & .295  & .197  & .096             & .096  & .279     \\
                          & Binary Relevance$^{\mathcal{T}}$     & .409    & .655  & .285   & .655  & .399   & .647   & .279  & .647  & .167  & \textbf{.295}    & .117  & .295      \\
                          & \textit{Dynamic} Triplet-loss$^{\mathcal{T}}$ & .449    & .447   & .408      & .539 & .404 & .353   & .382   & .513     & .559    & .166   & .344   & .631      \\
\midrule
\parbox[t]{2mm}{\multirow{5}{*}{\rotatebox[origin=c]{90}{XMTC}}}
                          & eXtremeText (Sigmoid)  & .659   & .649   & .426   & .713  & .742   & .704   & .494  & .793  & .439  & .212             & .333  & .521     \\
                          & eXtremeText (PLT)      & .644   & .636   & .403   & .689  & .714   & .679   & .464  & .756  & .465  & .206             & .327  & .532     \\
                          & NAPKINXC               & .698   & .687   & .426   & .764  & .800   & .748   & .495  & .864  & .548  & .253             & .409  & .626     \\
                          & XR-LINEAR              & .705   & .700   & .429   & .772  & .817   & .765   & .494  & .870  & .586  & \underline{.261} & .439  & .669     \\
                          & XR-TRANSFORMER$^{\mathcal{T}}$         & .683   & .673   & .416   & .747  & .801   & .750   & .488  & .856  & .554  & .245             & .405  & .633     \\
\midrule
\parbox[t]{2mm}{\multirow{4}{*}{\rotatebox[origin=c]{90}{Ours}}}    
                          & InfoNCE$^{\mathcal{T}}$       & .759   & .727   & .624      & .823     & .819   & .696   & .668      & .876     & .741   & .228   & \textbf{.515}      & \textbf{.871}     \\
                          &  $@-$balanced$^{\mathcal{T}}$         & \textbf{.843}   & \underline{.806}   & \underline{.666}      & \underline{.892}     & \underline{.889}   & \underline{.778}   & .711      & \underline{.927}     & .731   & .224   & .491      & .789     \\
                          & \hspace{2mm} \textit{w/o} auxiliary     & .714   & .689   & .579      & .791     & .817   & .697   & .648      & .88     & \textbf{.754}  & .233  & \underline{.509}      &  \underline{.816}     \\
                          & \hspace{2mm} \textit{w/o} Transformers    & .777   & .733   & .664      & .86     & .791   & .683   & \underline{.713} & .875     & .718   & .226   & .497      & .782  \\  
                          & Asymmetric$^{\mathcal{T}}$       & \underline{.841}   & \textbf{.806}   & \textbf{.677}      & \textbf{.892}     & \textbf{.903} & \textbf{.789}   & \textbf{.726}      & \textbf{.938}     & \underline{.745}   & .236   & .483      & .802    \\
\bottomrule
\end{tabular}
}
\end{table*}

\subsection{Metrics and Baselines}
\label{sec:baselines}
The following common metrics in literature are used: the micro-averaged $\{\textbf{P},\textbf{R},\textbf{F1}\}@k$ and mean reciprocal rank (\textbf{MRR})$@k$, which measures the relative ordering of a ranked list.

The following baselines are targeted: \textbf{Okapi BM25}, adjusted from~\citet{Husari_Al-Shaer_Ahmed_Chu_Niu_2017}. The BoW is augmented with $k$ closest terms from a security GloVe model, enhancing the BM25 retrieval capability. Here, \textit{query} represents the target text, and \textit{documents} refer to TTP descriptions. 

\textbf{Binary Relevance}, the vanilla multi-label learning approach, similar to~\citet{Li_Zheng_Liu_Yang_2019} for TTP mapping. It has the one side of the text matching architecture and learns a binary classifier for each label separately in a \textit{one-vs-all} manner. 

\textbf{Dynamic \textit{triplet}-loss}, a competitive baseline with a similar network architecture to ours, employs a \textit{triplet}-based loss~\cite{schroff2015facenet}. In contrast to the (empirically found) ineffective vanilla setting, we dynamically generate $k$-negative samples (akin to N-pairs loss~\cite{sohn2016improved}) to mimic the NCE mechanism.

In addition, we employ the following state-of-the-art (SoTA) models in XMTC as competitive baselines: \textbf{NAPKINXC~\cite{jasinska2020probabilistic}}, a method that generalized the Hierarchical Softmax, so-called Probabilistic Label Trees (PLT),  commonly used in XMTC literature. \textbf{XR-LINEAR~\cite{yu2022pecos}}, a model designed for very large output spaces, with~3 phases:  semantic label indexing (label clustering), matching (where the most relevant clusters are identified), and ranking (of labels in the matched clusters). 
\textbf{XR-TRANSFORMER~\cite{zhang2021fast}}, similar to XR-LINEAR, but with a transformer encoder. ~\textbf{exTremeText~\cite{wydmuch2018no}}, algorithm-wise relatively similar to NAPKINXC. 

\subsection{Experimental Setup}
We use the common security LM SecBERT\footnote{\url{https://github.com/jackaduma/SecBERT}} for the transformer-based models, and grid search determined the best hyperparameters for each model. The rich textual description\cref{mitrenote} of a TTP is selected for the textual profile. Except for XMTCs and BM25, all models are with the \textit{auxiliary} tasks. 

 
\textbf{Data Settings}. For the \textit{Procedure+} and TRAM datasets, each was \textit{stratified}-shuffled and split into training, validation and test sets with ratios of 72.5\%, 12.5\% and 15\%, respectively. The test sets remained fixed for reporting purposes. For training and validation, two modes were considered: \textit{separate} and \textit{combined}. In the former, the datasets are kept distinct, while in the latter, they were merged according to their respective splits. 

For the Expert dataset, we utilize a dedicated \textit{held-out} recall-focused test set, with~157 unique paragraph-level samples and~3.3 labels per sample on average. This carefully curated held-out set closely resembles paragraph-level text snippets in complete CTI reports, facilitating a comprehensive analysis of the entire report. 


\subsection{Results and Analysis}
Table~\ref{tab:results-all} presents the main experimental results. Overall, our proposed NCE-based models greatly outperform the baselines. Particularly, the \textit{asymmetric} loss-based model achieves the best performance across most metrics and datasets. We also observe substantial improvements of the two loss variants (i.e., $\alpha$-balanced and \textit{asymmetric}) over the vanilla InfoNCE. In addition, the models demonstrates a substantial improvement at the cutoff threshold $@1$ ($\sim$10\%) in comparison to $@3$ ($\sim$5\%). This supports the effectiveness of our \textit{matching} network in \textit{classification} settings.

The SoTA XMTC baselines perform considerably robust across the three datasets, among these XR-LINEAR perform best. Interestingly, XR-LINEAR demonstrates consistently higher performance than its related transformer-based counterpart (XR-TRANSFORMER), suggesting the challenges of the larger models in our low-resource settings. We also observe the subpar performance of the \textit{triplet}-loss approach, suggesting similar disadvantages in the low-resource settings. 

Across the datasets, the overall model performance declines from Procedure+ to TRAM and Expert, indicating varying complexities within each dataset. Notably, our performance yields compelling results in TRAM, well-surpassing methods commonly reported in related work, i.e., BM25 and Binary Relevance.


\subsection{Ablation Studies}

\textbf{Hierarchical Labeling}. We analyze the contributions of our \textit{hierarchical} modeling to the ranking performances. As shown in Table~\ref{tab:results-all}, in general, our joint learning with the \textit{auxiliary} task gives a notable performance boost in most scenarios. 
We report further in Table~\ref{tab:results-only-technique} the models' results in the \textit{technique}-level of the label hierarchy, where a sub-technique label is resolved to its technique. This is also a common practice in literature to streamline the complexity of the task. Overall, all models present substantial improvements in this setting. Interestingly, here the $\alpha$-balanced model, without the \textit{auxiliary} task, is the best performer on the Expert dataset. This is, nonetheless, understandable as the original hierarchical structure is semantically one level reduced in this case.

\textbf{Transformers}. We observe the positive contributions of SecBERT to the performance of all models in most cases. Nevertheless, without SecBERT (i.e., \textit{w/o} Transformers), our models are still very much on par with the strong XMTC baselines at $k=1$ and outperform them at  $k=3$, indicating the better ranking capability, specially on the \textit{Expert} dataset.

\textbf{Long Tail Analysis}.
Tables~\ref{tab:long-tail-tram} and~\ref{tab:long-tail-expert} provide an analysis on the models' performances on the classes of \textit{head} versus \textit{tail} frequency distributions visualized in Fig.~\ref{fig:ddist}. Overall, \textit{matching}-based approaches, with the inductive bias, 
are relatively robust, whereas the classification-based XMTC baselines suffer in the long tail.

\begin{table}[t]
\caption{Model performance on the \textit{head} vs. \textit{tail} parts of the TRAM dataset. \textit{Head} denotes more frequent TTPs ($>$ \textit{empirical} \textbf{7} samples in the \textit{training} split), whereas \textit{tail} are infrequent TTPs. All are trained in \textit{combined} mode. \textbf{Bold} denotes \textit{absolute} best performers.}
\label{tab:long-tail-tram}
\resizebox{\columnwidth}{!}{%
\begin{tabular}{cccclll}
\toprule
                 & \multicolumn{3}{c}{\textbf{TRAM} \textit{head (94.5\%)}} & \multicolumn{3}{c}{\textbf{TRAM} \textit{tail (5.5\%)}} \\ 
\cmidrule(r){2-4} \cmidrule(r){5-7} 
\textbf{Methods} & F1$@$1           & F1$@$3          & MRR$@$3           & F1$@$1              & F1$@$3             & MRR$@$3       \\
\midrule
BM25             & .195           & .112          & .21            & +118\%            & +99.1\%          & +108\%                   \\
NAPKINXC         & .624           & .458          & .752           & -36.9\%           & -27.1\%          & -30.2\% \\
XR-LINEAR        & .62            & .448          & .743           & -16.3\%           & -25.4\%          & -21.5\%           \\
$\alpha$-balanced       & .668           & \textbf{.548} & .841           & \textbf{-3.3}\%   & \textbf{-12.2\%} & \textbf{-8\%} \\
Asymmetric       & \textbf{.679}  & .547          & \textbf{.848}  & -4.9\%            & -14.3\%          & -10.4\%         \\
\bottomrule
\end{tabular}
}\setlength\itemsep{-1em}
\end{table}
\setlength\itemsep{-1em}
\begin{table}[t]
\caption{Model performance on the \textit{head} vs. \textit{tail} parts of the Expert dataset. Legend of Table~\ref{tab:long-tail-tram} applies.}
\label{tab:long-tail-expert}
\resizebox{\columnwidth}{!}{%
\begin{tabular}{cccclll}
\toprule
                 & \multicolumn{3}{c}{\textbf{Expert} \textit{head (56.5\%)}} & \multicolumn{3}{c}{\textbf{Expert} \textit{tail (43.5\%)}} \\ 
\cmidrule(r){2-4} \cmidrule(r){5-7} 
\textbf{Methods} & F1$@$1        & F1$@$3       & MRR$@$3       & F1$@$1        & F1$@$3       & MRR$@$3       \\
\midrule
BM25             & .071           & .107 & .188           & +26\%           & +28\%         & +18.6\%           \\
NAPKINXC         & .334           & .381 & .655           & -40.7\%           & -23.9\% & -16.6\%           \\
XR-LINEAR        & \textbf{.335}           & .407 & .676           &  -31.6\%          & -22.9\% & -14.5\%          \\
$\alpha$-balanced       & .302           & \textbf{.426} & .819           & -18.2\% & \textbf{-11.3\%}          & -2.9\% \\
Asymmetric       & .306          & .416          & \textbf{.831}           & \textbf{-18.9\%} & -12\% & \textbf{-2.9\%}          \\
\bottomrule
\end{tabular}
}\setlength\itemsep{-1em}
\end{table}
\setlength\itemsep{-1em}

\begin{figure}[ht]
  \centering
  \includegraphics[width=\linewidth]{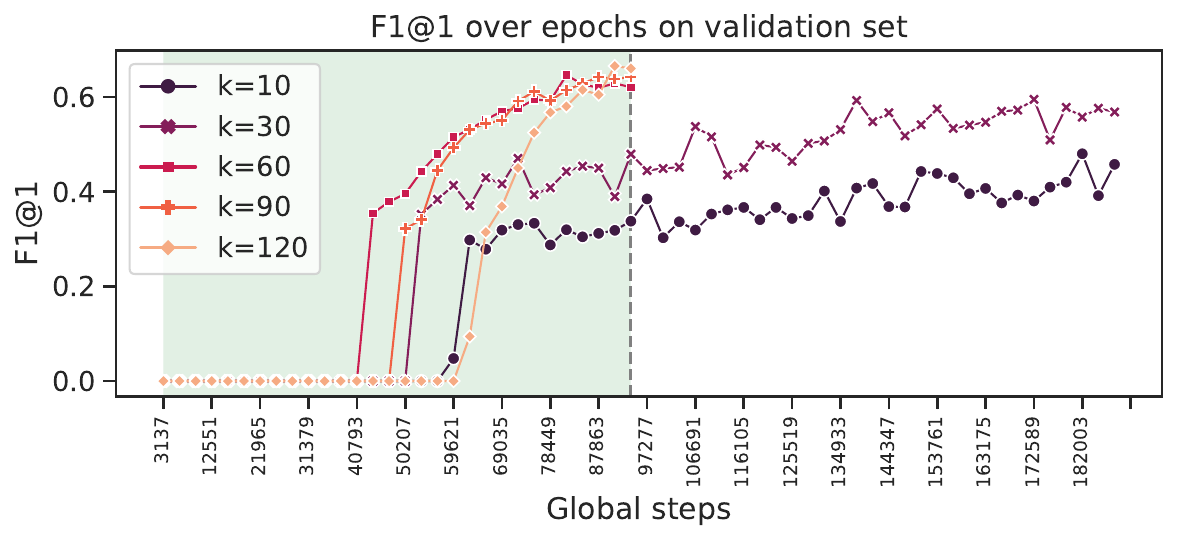}
  \hfill
   \centering
  \includegraphics[width=\linewidth]{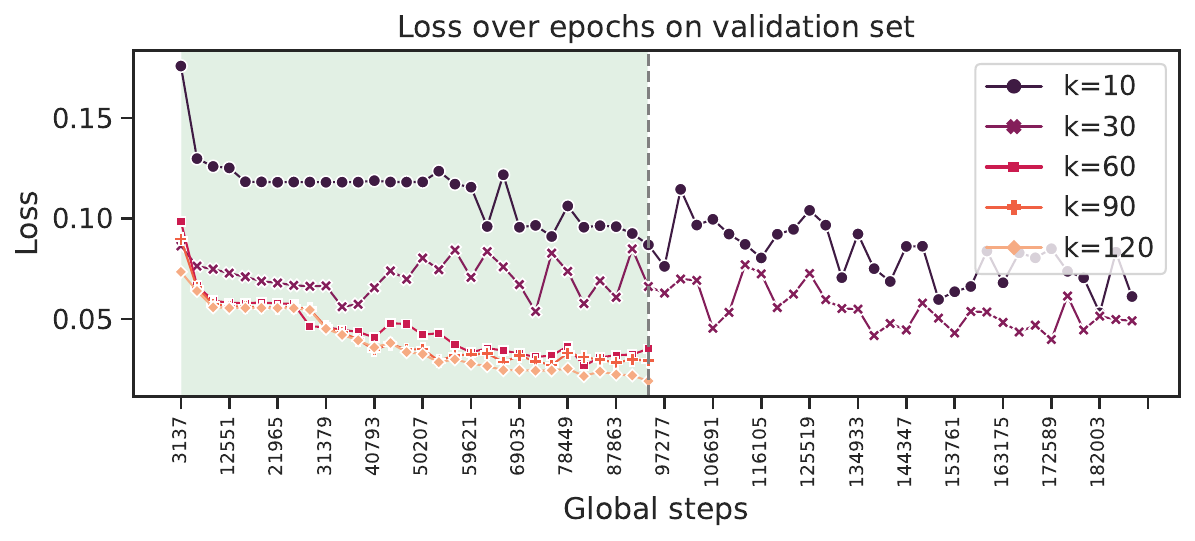}
  \caption{InfoNCE loss and f1$@$1 performance wrt. different number of negative samples. The network is without transformers. OOM for larger number of negative samples on an NVIDIA V100 32GB RAM.}
  \label{fig:f1-ks}
\end{figure}

\begin{figure}[ht]
  \centering
  \includegraphics[width=\linewidth]{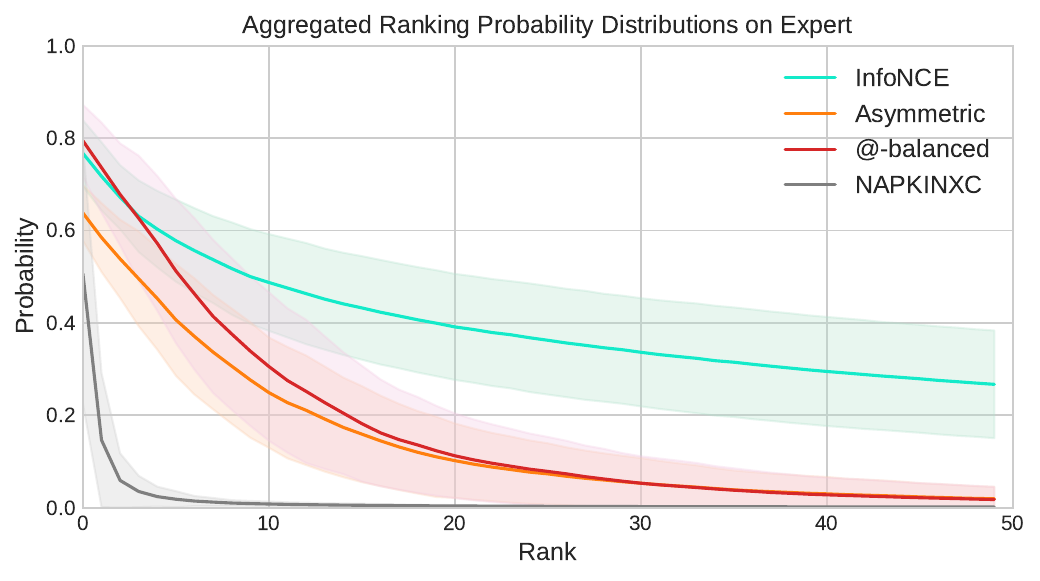}
  \hfill
   \centering
  \includegraphics[width=\linewidth]{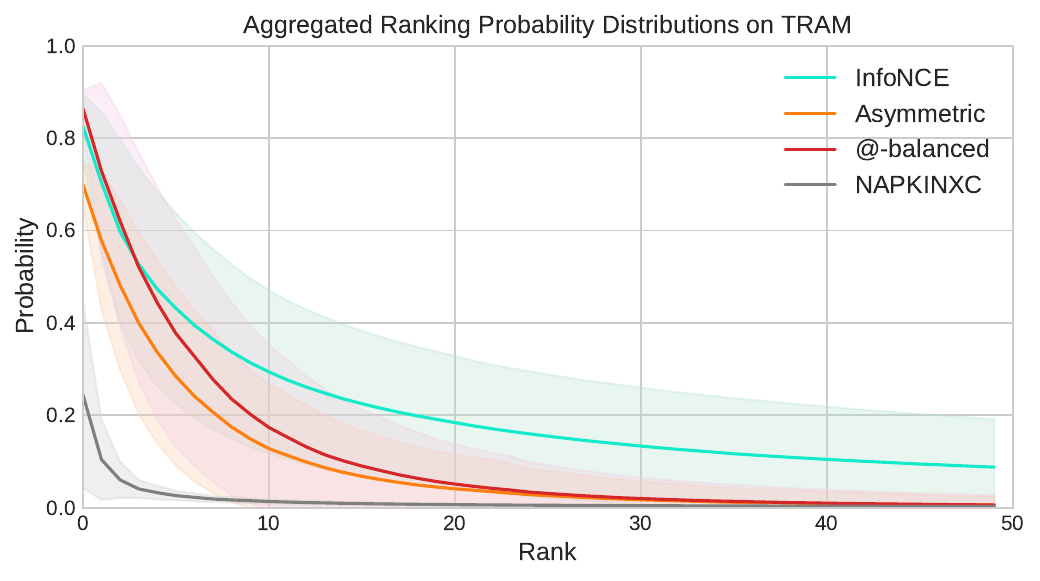}
  \caption{The aggregated probability distribution of the top-50 ranking on different models on the test splits of the TRAM (left) and Expert (right) datasets. While InfoNCE tends to allocate probabilities to labels in the long tail, $\alpha$-balanced and \textit{asymmetric} exhibit a more pronounced skewness in their distribution, resembling that of a pure classification model like NAPKINXC. The NCE-based models display a broader distribution at the head, indicating their inclination to assign comparable probabilities to multiple labels.}
  \label{fig:dist-tram}
\end{figure}

\textbf{Loss Analysis}.
In Fig.~\ref{fig:f1-ks}, we present additional analysis on the impact of the \textit{size} of negative samples. The results indicate that as the size increases, the model tends to converge faster and exhibit better performance. However, it appears that there are no additional benefits beyond a size of 60, which corresponds to 10\% of the label space.

A further analysis on the score distribution of the ranked lists are reported in Fig.~\ref{fig:dist-tram}. The details are provided in the caption for convenient reference.

\textbf{Expert Dataset}. 
To further examine the difficulties posed by the Expert Dataset, we present the outcomes of models trained on the training splits of \textit{Procedure+} and TRAM, evaluated on the entire Expert dataset. The results are showcased in Tables~\ref{tab:enex} and \ref{tab:enex-ot}. Overall, although all models exhibit reduced performance in this scenario, our models demonstrate superior generalization capability. Also, InfoNCE performs rather robustly in this setting, perhaps due to its stable nature to noisy input representation stemming from long-form text.

\begin{table}[htb]
\caption{Results on the \textit{entire} Expert dataset, trained on the training splits of \textit{Procedure+} and Tram. Bold denotes best-performer.}
\label{tab:enex}
\resizebox{\columnwidth}{!}{%
\begin{tabular}{cllllll}
\textbf{Methods} & \textbf{P$@$1}  & \textbf{R$@$1}  & \textbf{F1$@$3} & \textbf{MMR$@$3} & \textbf{F1$@$5} & \textbf{MRR$@$5} \\
\hline
TTPDrill (BM25)  & .311          & .166          & .226          & .364           & .207          & .375           \\
NAPKINXC         & .43           & .186          & .3            & .51            & .275          & .519           \\
XR-LINEAR        & .426          & .198          & .311          & .517           & .275          & .529           \\
\hline
InfoNCE          & \textbf{.489} & .208          & \textbf{.362} & \textbf{.564}  & \textbf{.339} & \textbf{.576}  \\
$\alpha$-balanced       & .443          & .195          & .328          & .528           & .324          & .543           \\
Asymmetric       & .484          & \textbf{.217} & .348          & .558           & .333          & .573          
\end{tabular}
}
\end{table}

\begin{table}[htb]
\caption{\textbf{Technique-level} results on the \textit{entire} Expert dataset. Legend in Table~\ref{tab:enex} applies.}
\label{tab:enex-ot}
\resizebox{\columnwidth}{!}{%
\begin{tabular}{cllllll}
\textbf{Methods} & \textbf{P$@$1}  & \textbf{R$@$1}  & \textbf{F1$@$3} & \textbf{MMR$@$3} & \textbf{F1$@$5} & \textbf{MRR$@$5} \\
\hline
TTPDrill (BM25)  & .369          & .202          & .283          & .437          & .267          & .449           \\
NAPKINXC         & .51           & .26          & .344            & .583            & .375          & .592           \\
XR-LINEAR        & .526          & .279 & .378          & .595           & .332          & .609           \\
\hline
InfoNCE          & \textbf{.556} & .286          & \textbf{.447} & \textbf{.621}  & \textbf{.432} & \textbf{.633}  \\
$\alpha$-balanced       & .506          & .273          & .428         & .594          & .429          & .604           \\
Asymmetric       & .543         & \textbf{.287} & .442          & .615           & .423 & .626         
\end{tabular}
}
\end{table}

\setlength\itemsep{-2em}
\section{Conclusion}
\setlength\itemsep{-1em}
We proposed a solution for the TTP mapping task that overcomes low-resource challenges in the security domain. The framework combines semantic label information with set-level $\alpha$-balancing, pair-level asymmetric focusing, and corpus-level sampling in a moderately sized label space. It separates collective candidate competition from pairwise influence and lets inexpensive partial rankings accumulate broad label-space exposure, while outperforming strong baselines. 

\clearpage
\section{Limitations}
Despite its label efficiency, our learning approach is not particularly efficient in terms of training. On average, it requires 24 hours for training on a machine equipped with a single NVIDIA-Tesla-V100 32 GB. Nonetheless, its training time is nearly comparable to the baselines employing Transformers.
Although our expert dataset closely aligns with the multi-label nature of the task and exhibits higher quality, it remains relatively limited in size, covering just one-third of the TTPs.

\section{Ethics Statement}
Our datasets are constructed from security threat reports published by security vendors, and copyrighted by their respective owners. We scraped and extracted textual contents from these public websites to build the datasets. The criteria for text selection was whether the text discusses TTPs.

Some source reports contain Personally Identifiable Information (PII) of report authors, threat actors (i.e., persons suspected of involvement in cybercrime) or victims (i.e., persons suspected of being targeted by cybercrime). In the text selection process, we screened for any PII and removed all uncovered instances. However, we cannot rule out the possibility that some PII might have been missed in that process. Thus, users wishing to use the data will need to accept our terms of use and report potential remaining instances of PII, which will be removed in a subsequent dataset update. Crucially, the potential remaining PII in the dataset has been originally published by the reports' authors and may still remain public on the original websites even after our dataset updates. 

The datasets have been annotated by security experts in our organization as part of their regular work under full-time employment contracts.

The language of the dataset is English, written by native and non-native speakers. 

We are not aware of any ethical implications stemming from the intended use of this dataset, i.e., TTP mapping.


\newpage

\bibliography{anthology,ttpm}
\bibliographystyle{acl_natbib}

\appendix
\section{The Task of TTP Mapping}
\label{sec:appendixa}

In the cybersecurity domain, one of the pillars of effective defense is \emph{Cyber Threat Intelligence} (CTI). 
An analog to military intelligence, CTI is tasked with collecting and organizing information on cyber threats such as \emph{threat actors}, their threat \emph{campaigns}, and malicious software, i.e., \emph{malware}. 
It can be traced back to ancient military-theoretical observations that understanding one's enemy is crucial to winning battles\footnote{
  ``If you know the enemy and know yourself, you need not fear the result of a hundred battles. 
  If you know yourself but not the enemy, for every victory gained you will also suffer a defeat. 
  If you know neither the enemy nor yourself, you will succumb in every battle.''~\cite{SunTzuArtWar}
}.

CTI describes cyber threats on three levels.
The \emph{strategic level} (e.g., periodicals on trends in the cyber risk landscape) describes high-level threat information and targets non-technical chief executives. 
The \emph{tactical level} (e.g., technical reports on individual threat actors) describes details on threat actors' behavior, for use by security managers. 
The lowest, \emph{operational level} (e.g., lists of malicious internet domains) describes specific threat indicators which may be directly used for defense (e.g., by blocking the offending domains).

While the value of CTI data is roughly proportional to its intelligence level, the difficulty of obtaining it is the opposite. 
Automated production only exists for operational CTI data, and higher levels require costly manual expert work. 
However, leading CTI community members regularly publish tactical and strategic CTI information in form of \emph{cybersecurity threat reports} -- digital documents with unstructured natural language text along tables and images, written using a domain-specific vocabulary, between hundreds and thousands of words long, and strongly interspersed with technical tokens such as URLs, hashes and similar.
Topically they cover profiles of major threat actors, summaries of threat campaigns, and malware analysis reports.
An illustrative excerpt is provided in Fig.~\ref{fig:attack_text_example1}.
Thus an opportunity arose for a fruitful application of NLP: automated extraction of high-value CTI data from natural language documents. 

In recent years, the NLP and cybersecurity communities have been engaged in exactly this direction. 
Early work targeted the operational level, extracting \emph{Indicators of Compromise} (IoCs), i.e., threat actor controlled internet domains, IP addresses, file hashes and URLs, from security articles, social media or forum posts.
Subsequent efforts targeted the tactical level, but the challenge there remains unsolved.

The tactical level characterizes adversaries' behavior, typically referred to as \emph{attack patterns}. 
Fig.~\ref{fig:attack_text_example1} illustrates, among others, (1) the use of a malicious email attachment to take control of a victim's system, and (2) encrypting data on the victim's system to extort money from the victim.
To facilitate reasoning about attack patterns, of which hundreds are documented, the community converged around a common framework called \emph{Tactics, Techniques and Procedures} (TTPs):

\begin{itemize}
  \item A \textbf{tactic} describes the purpose of the actor's behavior -- ``why?''. 
        For above examples, the tactics are \emph{taking control of the system} and \emph{financial gain}, respectively.
        Other typical adversarial tactics include \emph{reconnaissance}, \emph{establishing permanent presence}, \emph{command and control}, \emph{data theft}, etc.
        
  \item A \textbf{technique} describes the method used for the given purpose -- ``how?''. 
        In our case, those are \emph{malicious email attachment} and \emph{data encryption}.
        A technique may be assigned to several tactics if it achieves several purposes.
        Each tactic can be achieved using any of a range of different techniques. 
        Other typical techniques include \emph{collecting victim system information}, \emph{execution on system start}, \emph{encrypted communication}, \emph{password theft}, etc.

  \item Some ontologies also define a \textbf{subtechnique} as a specialized technique.
        A technique may be specialized by zero or more subtechniques.
        For example, the technique \emph{input capture} may have subtechniques \emph{keystroke capture} and \emph{screen capture}.
        
  \item A \textbf{procedure} describes the implementation details of a technique. 
        For example, the email attachment may be a \emph{malicious Excel file}, and the data encryption may be performed using \emph{a custom encryption algorithm}.
        Each technique can be implemented using any of potentially many different procedures.
\end{itemize}

Although others exist, MITRE ATT\&CK\footnote{
  \url{https://attack.mitre.org/}
}~\cite{MitreAttackPhilosophy} is the prevalent knowledge base and taxonomy of TTPs used in the literature.
The version 12.0 comprises~14 tactics, 196~techniques, 411~sub-techniques and thousands of procedures, continually curated by community experts.

Retrieval of TTPs from unstructured text is referred to as \emph{TTP mapping} in this work, although \emph{TTP mining/extraction} also occur in the literature. 
Crucially for TTP mining, threat reports very rarely name actors' TTPs explicitly. 
Instead, they establish a chronological narrative in terms of \emph{threat actions}, i.e., low-level actions taken by the threat actor. 
Some examples for threat actions from Fig.~\ref{fig:attack_text_example1} are \emph{botnet spreading}, \emph{use of phishing emails}, \emph{use of Visual Basic for malicious scripting}, \emph{use of Excel macros}, etc.
Not all threat actions are explicitly expressed in the text.
For example, although the term ``email'' is not mentioned, the use of phishing emails is inferred by domain experts because phishing means sending deceptive emails with malicious purposes, therefore sending emails is the technical implementation of phishing and it must have occurred. 

Thus, at a high level, TTP mapping from text is a~3-step process:
\begin{enumerate}
  \item Identification of individual threat actions from paragraphs or longer context
  \item Correlation of one or more identified threat actions into procedures
  \item Mapping of identified procedures into techniques and tactics.
\end{enumerate}


\section{Theoretical and Optimization Properties}
\label{apd:conv}
We state properties used to interpret the sampled losses and their first-order optimization. They do not imply convergence to a local or global minimum, nor establish statistical consistency of the learned ranking scores.

\textbf{$\alpha$-balanced gradient.}
For one sampled candidate set, let
\begin{equation}
\begin{aligned}
Z_{\gamma}
  &=\exp(g_{+})+\gamma\sum_{j=1}^{K}\exp(g_j^{-}),\\
\pi_{+}
  &=\frac{\exp(g_{+})}{Z_{\gamma}},\\
\pi_j^{-}
  &=\frac{\gamma\exp(g_j^{-})}{Z_{\gamma}}.
\end{aligned}
\end{equation}
The per-example loss in Equation~\ref{eq:infonce} is $\ell_{\alpha\text{-bal}}=-\log\pi_{+}$, and
\begin{equation}
\begin{aligned}
\nabla_{\theta}\ell_{\alpha\text{-bal}}
={}&-(1-\pi_{+})\nabla_{\theta}g_{+}\\
&+\sum_{j=1}^{K}\pi_j^{-}\nabla_{\theta}g_j^{-}.
\end{aligned}
\label{eq:balanced-gradient}
\end{equation}
Because $1-\pi_{+}=\sum_j\pi_j^{-}$, the assumption
$\|\nabla_{\theta}g_{\theta}(x,r(\ell))\|\leq B_g$ on the parameter region visited by training implies
\begin{equation}
\|\nabla_{\theta}\ell_{\alpha\text{-bal}}\|
\leq 2B_g.
\end{equation}
Unlike a multiplicative constant outside the logarithm, $\gamma$ enters $Z_{\gamma}$ and therefore changes the normalized weights and the gradient.

\textbf{Asymmetric-loss derivatives.}
Let $p=\sigma(g)$ and $p_m=(p-m)_{+}$. For the positive loss in Equation~\ref{eq:asym-pair},
\begin{equation}
\frac{\partial \ell^{(+)}}{\partial g}
=(1-p)^{\gamma_{+}}
\left[\gamma_{+}p\log p-(1-p)\right].
\label{eq:asym-pos-gradient}
\end{equation}
For the negative loss, define
$A_{-}(u)=u/(1-u)-\gamma_{-}\log(1-u)$. Then
\begin{equation}
\frac{\partial \ell^{(-)}}{\partial g}
=
\begin{cases}
0, & p\leq m,\\[3pt]
p(1-p)p_m^{\gamma_{-}-1}A_{-}(p_m),
& p>m.
\end{cases}
\label{eq:asym-neg-gradient}
\end{equation}
For numerical stability, the arguments of the logarithms can be clipped to $[\varepsilon,1-\varepsilon]$. On a bounded parameter region with bounded input representations and bounded score Jacobians, Equations~\ref{eq:asym-pos-gradient}--\ref{eq:asym-neg-gradient} are bounded. These are sufficient assumptions, not automatic properties of an unconstrained neural network.

\textbf{Conditional SGD guarantee.}
Let $F(\theta)=\mathbb{E}_{\xi}[\ell(\theta;\xi)]$ denote the expected training objective, where $\xi$ contains the sampled positive pair and candidate negatives. For the smooth $\alpha$-balanced loss, or for a differentiable approximation of the hard cutoff in Equation~\ref{eq:asym-pair}, assume that (i) $F$ is lower bounded and has an $L$-Lipschitz gradient on the parameter region visited by training; (ii) the stochastic gradient is unbiased, $\mathbb{E}[G_t\mid\theta_t]=\nabla F(\theta_t)$; (iii) its conditional variance is bounded; and (iv) the learning rates satisfy $\sum_t\epsilon_t=\infty$ and $\sum_t\epsilon_t^2<\infty$. Under these standard non-convex stochastic-optimization conditions, SGD yields
\begin{equation}
\lim_{T\rightarrow\infty}
\frac{\sum_{t=0}^{T-1}\epsilon_t\,
\mathbb{E}\|\nabla F(\theta_t)\|^2}
{\sum_{t=0}^{T-1}\epsilon_t}
=0.
\label{eq:sgd-stationarity}
\end{equation}
Equation~\ref{eq:sgd-stationarity} is a first-order stationarity guarantee in an averaged sense. It does not guarantee that the iterates converge to a particular local minimum. The hard cutoff is only piecewise smooth, so a generalized-gradient analysis would require additional assumptions.

\textbf{Statistical scope.}
Uniform $q_x$ has full support over the admissible candidate labels, but the sampled objectives are defined relative to this choice of $q_x$. Moreover, incomplete annotations can create false negatives, and both the $\alpha$-balanced weighting and asymmetric focusing modify classical NCE. We therefore use these objectives as ranking losses and do not claim classical NCE density-ratio consistency for them.

\section{Dataset Construction}
\label{sec:dataconst}

\textbf{Derived Procedure Examples.}
The dataset is created as a contextualized version of the original \textit{Procedure} examples. We search for evidential \textit{paragraph-level} text snippets in the references where the summary example is derived from. With this, the examples are contextualized and reflect the  true reporting style present in the references. The pre-processing steps are as follows:

\begin{itemize}
\item Each example-reference pair is indexed at the \textit{paragraph} level. Any paragraphs that are deemed (1) too short (less than 20 tokens), (2) too long (more than 300 tokens), or (3) have a Jaccard index with the example exceeding 0.9 (indicating near-\textit{duplicate}) are discarded.
\item  The remaining paragraphs are ranked based on their relevance to the example using a tailored BM25 retrieval model.
\item A maximum of two paragraphs that satisfy a carefully chosen global cut-off threshold are selected.
\item Additionally, we eliminate any potential near-duplicates to the TRAM and Expert datasets. 
\end{itemize}

We further assessed the dataset quality on a limited sample set consisting of~50 text snippets. Through this qualitative evaluation, the overall impression of the examined samples is largely positive.

\textbf{Expert Dataset.}

The Expert dataset comprises relevant text paragraphs from articles of reputable cybersecurity threat researchers, annotated by seasoned cybersecurity experts. 
The dataset was purposefully designed to closely mimic real-world scenarios, aiming to provide a practical and authentic setting for TTP extraction. Unlike datasets that primarily focus on individual sentences, our dataset encompasses entire paragraphs, and the annotations are inherently multi-label in nature. Rather than concentrating on isolated sentences, this dataset includes entire paragraphs that contain implicit mentions of TTPs, making the annotations inherently multi-label in nature.

The dataset was collected as follows:
\begin{enumerate}
    \item We scraped 30 thousand articles from the feeds of leading cyber threat research organizations, and heuristically filtered out irrelevant articles, which do not describe attacks related to malware, advanced persistent threats, or cyber threat campaigns.
    \item Further heuristics were applied to remove irrelevant paragraphs, i.e., we look for paragraphs which satisfy aforementioned length constraints, and contain at least~3 cybersecurity entities (e.g., malware, URL, etc.). The remaining relevant paragraphs were then randomly sub-sampled for annotation.
    \item The expert annotators were tasked with analyzing the paragraph and identifying TTPs. To assist them in this process, an in-house search engine, powered by the baseline retrieval model BM25, was employed. This search engine allowed the annotators to formulate queries based on the paragraph and retrieve relevant information to aid in their TTP selection.
    \item The annotators were instructed to only annotate explicit tactics and techniques in the given paragraph\footnote{An expert may comprehend from the text that it would be impossible to perform a discussed attack step without another tactic or technique, even if those dependencies were not explicitly written.}.
\end{enumerate}

Each annotated item, namely a text paragraph, undergoes evaluation by a single annotator.
We refrained from implementing extra quality control procedures, such as reviews or reaching consensus among annotators. To ensure quality, we engaged seasoned cybersecurity experts as annotators, rather than relying on crowd-sourced workers.

The choice of text paragraphs is biased by the described selection process towards high-quality writing from expert threat reports, and might not be representative of other writing styles, e.g., micro-blogging posts.


\textbf{Expert Dataset: Special Test Split.}
In the aforementioned process, it cannot be guaranteed that all annotations will be retrieved accurately due to the extensive task of re-formulating queries and reviewing the lengthy ranked list of TTPs generated by the relatively lower-performing BM25 model. Therefore, in order to enhance the recall of the test split, we substituted BM25 with our \textit{InfoNCE} model, which was trained on the train splits of the \textit{Procedure+} and Tram datasets. For every sample, we utilize a deep cut-off approach by selecting the top 20 entries, which are then assigned to annotators for further analysis. We continued to follow the same procedures as before.

In rare cases, relevant labels were missing from the top-20 predictions, but the annotators were not explicitly instructed to manually include those labels in the dataset. 
Thus the recall of the annotations is inherently imperfect, and the labels tend to be biased towards to the use of InfoNCE,
Nevertheless, based on the annotators' subjective assessment, the estimated annotation recall ranged from 95-100\%, indicating that this dataset deviates minimally from a perfect annotation.
Consequently, this split contains a markedly higher number of labels per sample compared to competing datasets., e.g., TRAM.


In conclusion, our Expert dataset, and particularly the test split, is of relatively small size, but is comprised of fully representative text paragraphs and has exemplary annotation precision and recall.

\section{Further Experimental Studies}

\subsection{Metrics}
The definitions of the used metrics in our experiments are reported below.

\textbf{P$@k$}. Given a ranked list of predicted labels for each sample, the micro precision of the top-k is defined as: $P@k = 1/k \sum_{i=1}^{k} 1_{y_{i}^{+}}(l_i)$, whereas $l_i$ is the i-th label in the ranking and $1_{y_{i}^{+}}$ is the indicator function.

\textbf{R$@k$}. Similarly, the micro recall of the top-k is defined as:  $R@k = 1/|Q| \sum_{i=1}^{k} 1_{y_{i}^{+}}(l_i)$, whereas $|Q|$ is the number of positive labels in the sample.

\textbf{F1$@k$}. The metric maintains the harmony between P$@k$ and R$@k$ of a given ranked list, and is calculated as $\frac{2 \cdot P@k \cdot R@k}{(P@k + R@k)}$.


\textbf{MRR$@k$}. The metric measures the relative ordering of a ranked list, with  RR is the inverse rank of the first relevant item in the top-k ranked list. Accordingly, MRR$@k$ is measured as follows. $MRR@k=1/S \sum_{i=1}^S 1/rank_i$, whereas $S$ is the number of samples.

\subsection{Training Procedure and Hyperparameters}
While InfoNCE and $\alpha$-balanced use the standard one-stage training procedure, to leverage the effectiveness of the \textit{asymmetric} loss, which performs optimally under stable gradient conditions, we adopt a two-step training procedure in our experiments. Initially, the model is trained using an $\alpha$-balanced loss. Once the training process reaches a stable state, we then introduce the \textit{asymmetric} loss.

We report the best hyperparameter sets for all models in Table~\ref{tab:hparams}.
For the XMTC baselines, the parameter ranges for the probabilistic-based tree construction (i.e., with Huffman or K-Means) are designed to closely resemble the structure of the ATT\&CK taxonomy. This resemblance is achievable thanks to its dot-separated naming convention, where the prefix represents the super technique.

\begin{table*}[t]
\caption{The default hyperparameters used in the experiments for each model.}
\label{tab:hparams}
\resizebox{\linewidth}{!}{%
\begin{tabular}{lll}
\toprule
                                          & \textbf{Models}       & \textbf{Hyperparams}  \\
\midrule
\multicolumn{1}{c}{\multirow{3}{*}{\rotatebox[origin=c]{90}{Ours}}} & $\alpha$-balanced            & \{cls-ratio: \{$\gamma$: 0.11\}\}                                                                                                                                \\
\multicolumn{1}{c}{}                      & InfoNCE            & \{cls-ratio: \{$\gamma$: 1.\}\}                                                                                                                                \\
\multicolumn{1}{c}{}                      & asymmetric            & \{$\gamma\_pos$:1, $\gamma\_neg$:3, cut-off: 0.1\}                                                                                                                 \\
\multicolumn{1}{c}{}                      & - base settings       & \begin{tabular}[c]{@{}l@{}}\{learning\_rate: 1e-3, auxiliary\_task: \{$\alpha$: 0.6, $\beta$: 0.4\}, batch\_size:{[}2,\textbf{4},8{]},\\  negative\_samples:{[}\textbf{30},60{]}\, sampling\_method: \textit{random}\}\end{tabular} 
                \\
                \multicolumn{1}{c}{}                      & - auxiliary            & \{$\alpha$: 0.6, $\beta$: 0.4\}                                                                                                                 \\
\midrule
                                          & Dynamic Triplet Loss     & \begin{tabular}[c]{@{}l@{}}\{cls-ratio: \{$\gamma$: 0.11\}\, learning\_rate: 1e-3, auxiliary\_task: \{$\alpha$: 0.6, $\beta$: 0.4\}, batch\_size:{[}2,\textbf{4},8{]},\\  negative\_samples:{[}\textbf{30},60{]}\, sampling\_method: \textit{random}\}\end{tabular}  \\
\midrule                                          
                                          & NAPKINXC              & \begin{tabular}[c]{@{}l@{}}\{model: PLT, tree\_type: \{``hierarchicalKmeans'', ``huffman''\}, \\ arity:\{2,10, 20\}, max\_leaves:\{10, 20\}, kmeans\_eps=0.0001, \\ kmeans\_balanced=\{True, False\}\}\end{tabular} \\
                                          & XR-LINEAR             & \{mode: ``full-model'', ranker\_level: 1, nr\_splits: 16\}                                                                                                           \\
                                          & XR-TRANSFORMER        & \begin{tabular}[c]{@{}l@{}}\{mode: ``full-model'', negative\_sampling: {[}``tfn'', ``man''{]},  \\ ,  do\_fine\_tune: True, only\_encoder: False\}\end{tabular}          \\
                                          & ExtremeText + Sigmoid & \{loss: sigmoid, neg: {[}0, 100{]}, tfidfWeights: True\}                                                                                                           \\
                                          & ExtremeText + PLT     & \{loss: ``plt'', neg: {[}0, 40{]}, tree\_type: \{``hierarchicalKmeans'', ``huffman''\}, tfidfWeights: True\} \\
\bottomrule
\end{tabular}
}
\end{table*}

\subsection{Qualitative Studies}
In this section, we provide a series of illustrative examples (see ~\Cref{tab:ex1,tab:ex2,tab:ex3}) to qualitatively showcase the practical efficacy of our methodology in addressing the compound TTP-Mapping task. We relate our results with the established LLMs, such as ChatGPT 4~\footnote{While being extensively studied, we opt to exclude its results in our experiments due to the \textit{objective} prompt-sensitive performance limitations.}, which serve as a reference to the overall intricacy of this task. 

For the setup of ChatGPT, for each text, we create a \textit{prompt} in the following format: \textit{What MITRE ATT\&CK techniques (TTPs) are explicitly and implicitly mentioned in the following text: [..]}.  In general, the responses provided by Chat-GPT are remarkable and somewhat accurate in certain instances. However, it is evident that the answers primarily consist of high-level information (sometimes hallucinatory), with a lack of granularity that makes it useful, e.g., for precise modeling of the attack steps.

We provide further a \textbf{full report analysis} of a threat report released by Mandiant (see \href{https://web.archive.org/web/20230316150631/https://www.mandiant.com/resources/blog/apt41-initiates-global-intrusion-campaign-using-multiple-exploits}{Wayback machine}). Each paragraph in the report is processed by our model and finally techniques were assigned to the \textit{tactic bins} of the MITRE ATT\&CK matrix~\footnote{\href{https://attack.mitre.org/matrices/enterprise/}{\nolinkurl{https://attack.mitre.org/}}}, based on a simple assignment algorithm, with two constraints (1) maximize total relevant score (of each TTP) in the bins and (2) maximize total number of TTP-occurrences in the bins (i.e., a TTP can occur in more than one paragraph). Further details are in Table~\ref{tab:apt41}.

\begin{table*}[htp]
\caption{A \textbf{full report analysis} of the Mandiant threat report (see \href{https://web.archive.org/web/20230316150631/https://www.mandiant.com/resources/blog/apt41-initiates-global-intrusion-campaign-using-multiple-exploits}{Wayback machine}). We compare our results with the list of TTPs explicitly provided by the same report, Appendix section, with that, we achieve 90\% recall, missing only one technique (Non-Standard Port). All the extracted TTPs from the model are further examined and confirmed correct by our security experts.}
\small
    \centering
    \begin{tabular}{lp{8cm}}
        \hline
        \textbf{Tactics} & \textbf{Techniques} \\
        \hline
        Reconnaissance& 
            \begin{itemize}[nosep, after=\strut]
                \item Vulnerability Scanning (T1595.002)
            \end{itemize} \\
        Resource Development& 
            \begin{itemize}[nosep, after=\strut]
                \item Vulnerabilities (T1588.006)
                \item Exploits (T1588.005)
            \end{itemize} \\
        Initial Access & 
            \begin{itemize}[nosep, after=\strut]
                \item External Remote Services (T1133)
                \item Exploit Public-Facing Application (T1190)
            \end{itemize} \\
        Execution & 
            \begin{itemize}[nosep, after=\strut]
                \item Windows Command Shell (T1059.003)
                \item Exploitation for Client Execution (T1203)
            \end{itemize} \\
        Persistence & 
            \begin{itemize}[nosep, after=\strut]
                \item BITS Jobs (T1197)
                \item Windows Service (T1543.003)
            \end{itemize} \\
        Privilege Escalation& 
            \begin{itemize}[nosep, after=\strut]
                \item Process Hollowing (T1055.012)
                \item Exploitation for Privilege Escalation (T1068)
            \end{itemize} \\
        Defense Evasion& 
            \begin{itemize}[nosep, after=\strut]
                \item Obfuscated Files or Information (T1027)
                \item Deobfuscate/Decode Files or Information (T1140)
            \end{itemize} \\
        Discovery& 
            \begin{itemize}[nosep, after=\strut]
                \item System Information Discovery (T1082)
                \item File and Directory Discovery (T1083)
            \end{itemize} \\
        Lateral Movement & 
            \begin{itemize}[nosep, after=\strut]
                \item Exploitation of Remote Services (T1210)
                \item Lateral Tool Transfer (T1570)
            \end{itemize} \\
        Command and Control & 
            \begin{itemize}[nosep, after=\strut]
                \item Ingress Tool Transfer (T1105)
                \item File Transfer Protocols (T1071.002)
            \end{itemize} \\
        \hline
    \end{tabular}
    \label{tab:apt41}
\end{table*}

\begin{table*}[t]
\caption{Example 1: A threat report text is analyzed by ChatGPT and $\alpha$-balanced. Here, an important technique is the \textit{Masquerading}, described by MITRE as: ``\textit{Adversaries may attempt to manipulate features of their artifacts to \textbf{make them appear legitimate} or benign to users and/or security tools. Masquerading occurs when the name or location of an object, legitimate or malicious, is manipulated or abused for the sake of evading defenses and observation. This may include manipulating file metadata, tricking users into misidentifying the file type, and giving legitimate task or service names.}`` This technique is missed by ChatGPT, identified by our model, and is ranked higher than explicit TTPs i.e., Powershell or Visual Basic.}
\small
\begin{tabular}{c p{0.8\textwidth}}
   \raggedright
Text& \textit{After the .NET PE file has been run, we observed the same behavior as the above QUADAGENT sample of dropping a PowerShell script with the filename SystemDiskClean.ps1 alongside a VBScript file with the same name. The C2 techniques remained identical, with the only change being the server which became cpuproc{[}.{]}com.\}} \\
ChatGPT& \que{\small Based on the provided text, we can identify several MITRE ATT\&CK techniques, both explicitly and implicitly: 
\begin{itemize}
    \item \textbf{Execution} (T1059): The text mentions the execution of a .NET PE file, indicating the execution technique.
    \item \textbf{Scripting} (T1064): The dropping of a PowerShell script (SystemDiskClean.ps1) and a VBScript file suggests the use of scripting techniques.
    \item \textbf{Command and Scripting Interpreter} (T1059.003): Specifically, the use of PowerShell in dropping the SystemDiskClean.ps1 file implies the use of command and scripting interpreters.
    \item \textbf{Deobfuscate/Decode Files or Information} (T1140): The use of a .NET PE file suggests potential obfuscation or encoding that may need to be deobfuscated or decoded.
    \item \textbf{Command and Control} (T1043): The mention of a change in the C2 server (cpuproc{[}.{]}com) indicates the use of a different command and control infrastructure.
\end{itemize}} \\
$\alpha$-balanced& 
\ans{\small The top-n ranking is:
\begin{enumerate}
    \item \textbf{Masquerading} (T1036), score: 0.68
    \item \textbf{PowerShell} (T1059.001),  score: 0.58
    \item \textbf{Visual Basic} (T1059.005), score: 0.52 
    \item \textbf{Obfuscated Files or Information} (T1027), score: 0.47 
    \end{enumerate}}
\end{tabular}
\label{tab:ex1}
\end{table*}

\begin{table*}[t]
\caption{Example 2: A threat report text is analyzed by ChatGPT and $\alpha$-balanced. Here, ChatGPT provides a correct justification on the use of the common HTTP port 443. This is a false positive from our model, as a simple text matching fails to \textbf{reason} whether a port number is common or uncommon. Our model however, is still able to identify the important technique \textit{Ingress Tool Transfer}, which described by MITRE as: ``\textit{Adversaries may transfer tools or other files from an external system into a compromised environment. Tools or files may be copied from an external adversary-controlled system to the victim network through the command and control channel or through alternate protocols such as ftp. Once present, adversaries may also transfer/spread tools between victim devices within a compromised environment.}''} 
\small
\begin{tabular}{c p{0.8\textwidth}}
   \raggedright
Text& \textit{The Meterpreter downloader ‘TzGG’ was configured to communicate with 91.208.184[.]78 over port 443 to download the shellcode (MD5: 659bd19b562059f3f0cc978e15624fd9) for Cobalt Strike BEACON (trial-version).} \\
ChatGPT& \que{\small Based on the provided text, we can identify several MITRE ATT\&CK techniques, both explicitly and implicitly: 
\begin{itemize}
\item \textbf{Data Obfuscation} (T1001): The use of the term "shellcode" implies the presence of binary code that is typically obfuscated and designed to be directly executed.
\item \textbf{Standard Application Layer Protocol} (T1071): The communication with IP address 91.208.184[.]78 over port 443 suggests the use of HTTPS, which is a standard application layer protocol.
\item \textbf{Command and Control} (T1043): The Meterpreter downloader 'TzGG' communicating with an external server over port 443 indicates a command and control channel.
\item \textbf{Connection Proxy} (T1090.002): The use of port 443 may involve the exploitation of the encrypted communication channel often used for securing web traffic, serving as a form of connection proxy.
\item \textbf{Impersonation of Legitimate Entity} (T1090.002): The mention of Cobalt Strike BEACON (trial-version) suggests the potential for attackers to impersonate legitimate entities or tools.
\item \textbf{Application Layer Protocol} (T1071.001): The use of port 443 for communication is specifically associated with the HTTPS protocol at the application layer.
\item \textbf{Download Evil Content} (T1105): The Meterpreter downloader 'TzGG' is configured to download shellcode for Cobalt Strike BEACON, indicating the download of malicious content.
\item \textbf{Remote Access Tools} (T1219): Cobalt Strike BEACON is a well-known remote access tool often used by adversaries for post-exploitation activities.
\end{itemize}} \\
$\alpha$-balanced& 
\ans{\small The top-n ranking is:
\begin{enumerate}
    \item \textbf{Non-Standard Port} (T1105), score: 0.76
    \item \textbf{Ingress Tool Transfer} (T1059.001),  score: 0.71
    \item \textbf{Application Layer Protocol} (T1071), score: 0.59 
    \item \textbf{Web Protocols} (T1071.001), score: 0.57
    \end{enumerate}}
\end{tabular}
\label{tab:ex2}
\end{table*}

\begin{table*}[t]
\caption{Example 3: A threat report text is analyzed by ChatGPT and $\alpha$-balanced. Here interestingly, both answers show certain consensus. Our model is still able to match to low-level techniques i.e.,  \textit{Symmetric Cryptography}, derived from the use of RC4 for encryption.} 
\small
\begin{tabular}{c p{0.8\textwidth}}
   \raggedright
Text& \textit{Unlike previously identified RedDelta PlugX samples, this one uses a legitimate Avast Proxy executable for the second DLL side-loading phase rather than the legitimate Adobe executable used in previously analyzed samples. When loaded into memory, the PlugX payload uses www.systeminfor[.]com for command and control — the same domain used across the Catholic church-themed PlugX samples. Across all four newly analyzed samples, the first stage DLL loader files share an identical, uncommon import hash and rich header with previously observed first stage RedDelta PlugX DLLs. The loaded PlugX payload also matches the custom RedDelta PlugX variant previously described, using RC4 encryption for C2 communications with the same hardcoded RC4 passphrase and configuration block decoding function.} \\
ChatGPT& \que{\small Based on the provided text, we can identify several MITRE ATT\&CK techniques, both explicitly and implicitly: 
\begin{itemize}
\item \textbf{DLL Side-Loading} (T1073.002): The mention of using a legitimate Avast Proxy executable for the second DLL side-loading phase indicates the use of DLL side-loading, a technique where a malicious DLL is loaded into a legitimate process.

\item \textbf{Masquerading} (T1036): The use of a legitimate Avast Proxy executable for the second DLL side-loading phase is a form of masquerading or disguising the malicious activity to appear legitimate.

\item \textbf{Command and Control} (T1043): The PlugX payload uses the domain www.systeminfor[.]com for command and control, indicating communication with an external server.

\item \textbf{Standard Application Layer Protocol} (T1071): The use of a domain for command and control suggests the use of standard application layer protocols for communication.

\item \textbf{Data Obfuscation} (T1001): The use of RC4 encryption for C2 communications indicates the obfuscation of communication traffic.

\item \textbf{RC4} (T1027): The mention of using RC4 encryption for C2 communications specifically refers to the use of the RC4 symmetric encryption algorithm.

\item \textbf{Data Encoding} (T1132): The use of RC4 encryption can be considered a form of data encoding for communication between the implant and the C2 server.

\item \textbf{Encoded/Encrypted Files or Information} (T1022): The RC4 encryption is used for encrypting communications between the PlugX payload and the C2 server, securing the information exchanged.

\item \textbf{Registry Run Keys / Startup Folder} (T1060): While not explicitly mentioned in the provided text, the persistence mechanism used by PlugX (loading into memory) often involves leveraging registry run keys or startup folders.
\end{itemize}} \\
$\alpha$-balanced& 
\ans{\small The top-n ranking is:
\begin{enumerate}
    \item \textbf{DLL Side-Loading} (T1574.002), score: 0.81
    \item \textbf{Obfuscated Files or Information} (T1027),  score: 0.56
    \item \textbf{DLL Search Order Hijacking} (T1574.001), score: 0.52 
    \item \textbf{Encrypted Channel} (T1573), score: 0.49
    \item \textbf{Symmetric Cryptography} (T1573.001), score: 045
    \item \textbf{Deobfuscate/Decode Files or Information} (T1140), score: 0.32
    \item \textbf{Masquerading} (T1036), score: 0.32
    \item \textbf{Registry Run Keys / Startup Folder} (T1547.001), score: 0.31
    \end{enumerate}}
\end{tabular}
\label{tab:ex3}
\end{table*}




\end{document}